\documentclass{article}

\usepackage{microtype}
\usepackage{graphicx}
\usepackage{subcaption}
\usepackage[table]{xcolor}
\usepackage{booktabs} 

\usepackage{hyperref}

\usepackage[preprint]{icml2026}

\usepackage{tabularx}
\usepackage{multirow}
\usepackage{makecell}
\usepackage{booktabs}
\usepackage{arydshln}

\usepackage{amsmath}
\usepackage{amssymb}
\usepackage{mathtools}
\usepackage{amsthm}
\usepackage{mathrsfs}
\usepackage{xcolor}
\usepackage[table]{xcolor}
\usepackage{colortbl}
\usepackage{array}
\usepackage[capitalize,noabbrev]{cleveref}

\theoremstyle{plain}

\theoremstyle{definition}

\theoremstyle{remark}

\usepackage[textsize=tiny]{todonotes}

\icmltitlerunning{}

\begin{document}

\twocolumn[
  \icmltitle{Quantum-Inspired Fine-Tuning for 
    Few-Shot AIGC Detection \\ via Phase-Structured Reparameterization}



  \icmlsetsymbol{equal}{*}

  \begin{icmlauthorlist}
    \icmlauthor{Kaiyang Xing}{ahu}
    \icmlauthor{Han Fang}{sna}
    \icmlauthor{Zhaoyun Chen}{iai}
    \icmlauthor{Zhonghui Li}{ahu}
    \icmlauthor{Yang Yang}{ahu,iai}
    \icmlauthor{Weiming Zhang}{stcu}
    \icmlauthor{Guoping Guo}{stcu}
  \end{icmlauthorlist}

  \icmlaffiliation{ahu}{Anhui University, Hefei, China}
  \icmlaffiliation{iai}{Institute of Artificial Intelligence, Hefei Comprehensive National Science Center, Hefei, China}
  \icmlaffiliation{sna}{National University of Singapore, Singapore, Singapore}
  \icmlaffiliation{stcu}{University of Science and Technology of China, Hefei, China}
  
  \icmlcorrespondingauthor{Han Fang}{fanghan@nus.edu.sg}
  \icmlcorrespondingauthor{Yang Yang}{sky\_yang@ahu.edu.cn}

  \icmlkeywords{Machine Learning, ICML}

  \vskip 0.3in
]



\printAffiliationsAndNotice{}  

\begin{abstract}

Recent studies show that quantum neural networks (QNNs) generalize well in few-shot regimes. To extend this advantage to large-scale tasks, we propose Q-LoRA, a quantum-enhanced fine-tuning scheme that integrates lightweight QNNs into the low-rank adaptation (LoRA) adapter. Applied to AI-generated content (AIGC) detection, Q-LoRA consistently outperforms standard LoRA under few-shot settings. We analyze the source of this improvement and identify two possible structural inductive biases from QNNs: (i) phase-aware representations, which encode richer information across orthogonal amplitude–phase components, and (ii) norm-constrained transformations, which stabilize optimization via inherent orthogonality. However, Q-LoRA incurs non-trivial overhead due to quantum simulation. Motivated by our analysis, we further introduce H-LoRA, a fully classical variant that applies the Hilbert transform within the LoRA adapter to retain similar phase structure and constraints. Experiments on few-shot AIGC detection show that both Q-LoRA and H-LoRA outperform standard LoRA by over 5\% accuracy, with H-LoRA achieving comparable accuracy at significantly lower cost in this task.

\end{abstract}

\section{Introduction}


Quantum machine learning has shown great promise due to the unique computational properties of quantum systems. Among its key models, quantum neural networks (QNNs) have been theoretically and empirically suggested to possess a strong generalization ability in few-shot classification regimes \cite{caro2022generalization}. This advantage is often attributed to the geometric structure of Hilbert space and the norm-preserving nature of quantum operations, which together constrain the model’s effective capacity and reduce overfitting in low-data settings. However, these findings remain limited to toy-scale architectures. This raises a natural question: \textbf{Can such sample-efficient inductive biases scale to large models and real-world tasks?}

\begin{figure}[t!]
\centering
\begin{subfigure}{0.48\linewidth}
  \centering
  \includegraphics[width=\linewidth]{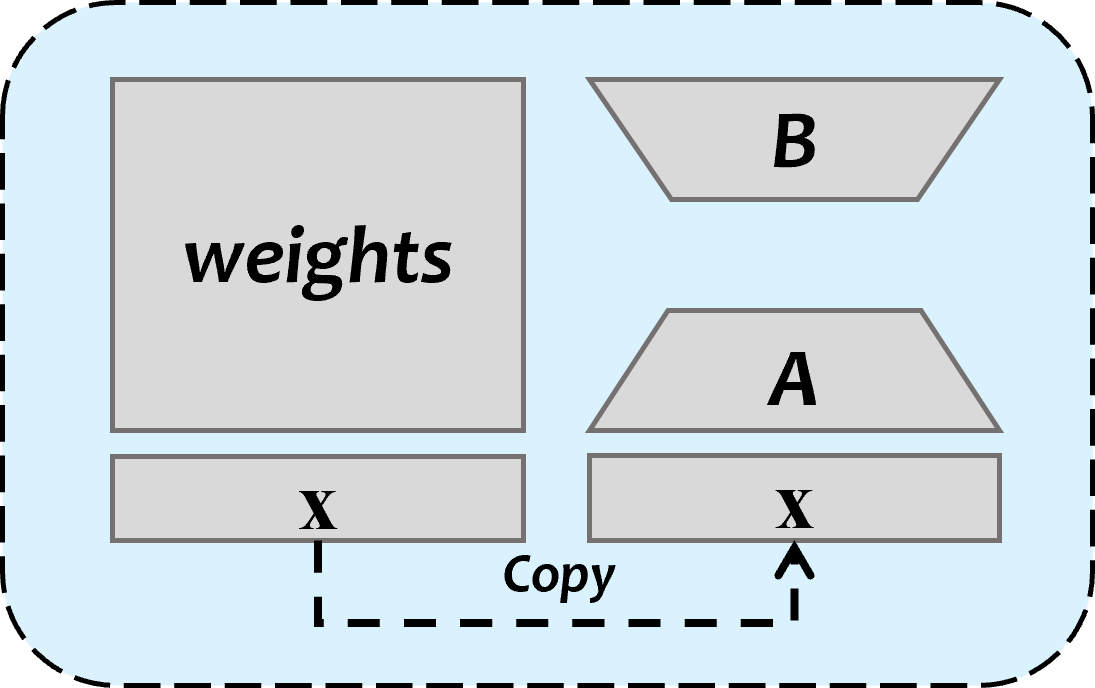}
  \caption{LoRA framework}
  \label{fig:LoRA}
\end{subfigure}\hfill
\begin{subfigure}{0.48\linewidth}
  \centering
  \includegraphics[width=\linewidth]{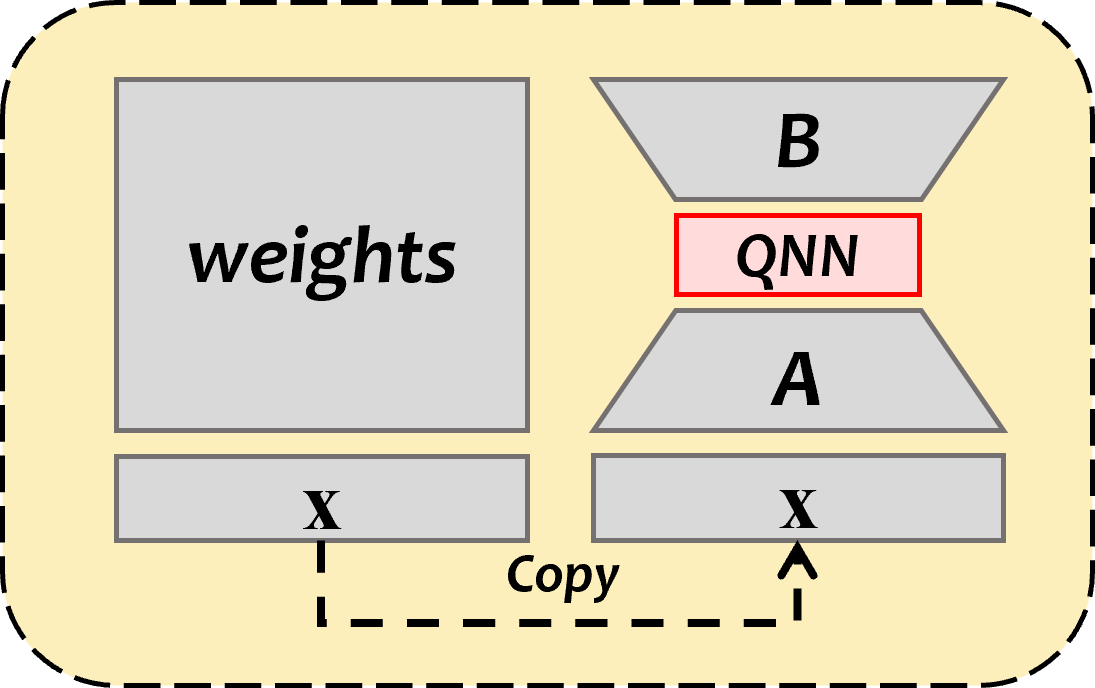}
  \caption{Q-LoRA framework}
  \label{fig:QLoRA}
\end{subfigure}

\begin{subfigure}{\linewidth}   
  \centering
  \includegraphics[width=\linewidth]{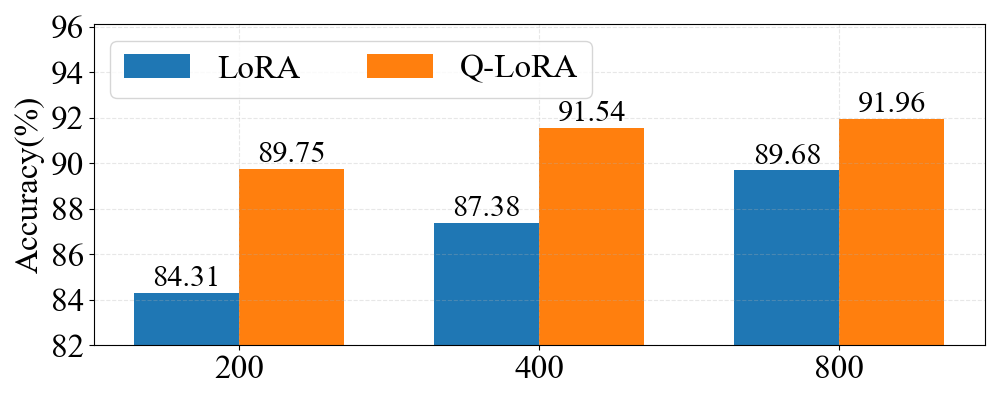}
  \caption{Detection accuracy of LoRA and Q-LoRA models with different training samples}
  \label{fig:lora_Q-lora_samples}
\end{subfigure}
\caption{Method of LoRA and Q-LoRA applied to the CLIP model, and their detection accuracy for AI-generated image detection tasks under different training samples.}
\end{figure}


To explore whether quantum generalization benefits can scale to real-world tasks, we propose Q-LoRA, a quantum-inspired fine-tuning scheme that integrates a lightweight QNN into the Low-Rank Adaptation (LoRA) \cite{hu2022lora} framework, which is a widely used parameter-efficient method for adapting large pre-trained models. Specifically, we instantiate Q-LoRA on top of CLIP \cite{radford2021clip}, freezing the backbone and injecting the QNN into the LoRA bottleneck. A shallow classification head is then trained to detect AI-generated content (AIGC).

As shown in Figure \ref{fig:LoRA} and \ref{fig:QLoRA}, the architectural modifications are minimal. Yet, as Figure \ref{fig:lora_Q-lora_samples} demonstrates, Q-LoRA consistently outperforms standard LoRA across varying few-shot sample sizes, with particularly notable gains in low-data regimes. These results suggest that the generalization advantage of QNNs, previously observed only in small-scale models, can be effectively transferred to large-scale fine-tuning tasks via quantum-inspired design. 



This performance improvement motivates a deeper inquiry into its underlying cause. However, deploying Q-LoRA in practice reveals a major limitation: substantial computational overhead due to quantum simulation. For instance, under a 200-sample training setting, Q-LoRA requires over 30 minutes per training epoch, compared to just a few seconds for standard LoRA. This pronounced efficiency gap calls into question the practicality of QNN-based methods for scalable fine-tuning.

To address this limitation while retaining the benefits of Q-LoRA, we turn to a more principled understanding: what exactly enables QNNs to generalize better? We hypothesize that the gains stem not from quantum mechanics per se, but from structural properties of QNNs that guide feature extraction and transformation. Specifically, we identify two key inductive biases: (i) phase-aware representations, where information is encoded across orthogonal amplitude–phase components, and (ii) norm-constrained transformations, which promote smoother optimization by limiting updates to orthogonal subspaces. These insights suggest that it may be possible to emulate Q-LoRA’s benefits in a fully classical setting, thereby avoiding the cost of quantum simulation while preserving its performance gains.

Guided by the above analysis, we propose H-LoRA, a fully classical variant that retains the core inductive biases of QNNs without relying on quantum computation. Concretely, H-LoRA applies the Hilbert transform to the feature stream within LoRA, generating analytic signals that couple real and imaginary components into orthogonal amplitude–phase representations. This operation inherently captures both structural advantages identified earlier:
(i) it introduces phase-aware encoding by enriching features with quadrature components;
(ii) it enforces norm-constrained transformations by maintaining fixed orthogonal coupling, thereby regularizing the optimization geometry. Experiments show that H-LoRA consistently matches or surpasses Q-LoRA in few-shot AIGC detection while reducing time consumption.

The core contributions of this paper are as follows:


\vspace{-0.5em}
\begin{itemize}
\item We empirically verify the generalization advantage of QNNs in large-scale few-shot tasks by introducing Q-LoRA, which integrates a lightweight QNN into the LoRA fine-tuning framework and achieves consistent gains on AIGC detection.
\item We conduct an in-depth analysis of where this advantage comes from, and distill two structural inductive biases: phase-aware encoding and norm-constrained transformations, which inspire the classical surrogate H-LoRA using Hilbert transform.
\item Extensive experiments show that H-LoRA and Q-LoRA both outperform LoRA. H-LoRA matches or surpasses performance to Q-LoRA while reducing time consumption. And validating the effectiveness of phase-structured inductive bias for few-shot adaptation in large models.
\end{itemize}

\section{Related Work}
\subsection{Low-Rank Adaptation}

The rapid scaling of large pre-trained models has motivated extensive research on parameter-efficient fine-tuning methods, which aim to approach the performance of full fine-tuning while substantially reducing computational and storage costs. Among these methods, LoRA is based on the observation that task-specific updates often lie in a low-rank subspace, and freezes the backbone while optimizing low-rank update matrices. This design significantly reduces the number of trainable parameters and often matches or even surpasses full fine-tuning performance with minimal inference overhead.

Despite its effectiveness, standard LoRA can suffer from suboptimal parameter utilization and limited generalization, especially in low-data regimes. Several extensions address these limitations by improving rank allocation or parameter efficiency, such as AdaLoRA, which dynamically redistributes rank budgets based on parameter importance \cite{zhang2023adalora}, and NoRM, which removes redundant components from trained LoRA adapters \cite{jiang2025norm}. More recently, Q-LoRA introduces quantum-inspired inductive biases into the LoRA framework via hybrid quantum–classical architectures, achieving improved generalization in few-shot settings \cite{kong2025quantumlora}.

\subsection{Hybrid Quantum-Classical Neural Network}




To explore the potential of quantum computing in machine learning and overcome the limitations of current quantum hardware, researchers have proposed quantum-classical hybrid models. The core idea is to delegate parts of the computation that are sensitive to quantum characteristics or benefit from high-dimensional Hilbert space processing to quantum units or simulators, while handling the remaining parts with efficient classical computers. Hybrid architectures integrating quantum circuits as trainable layers in neural networks have shown potential in fields such as healthcare, AI security, and image processing \cite{gordienko2025hnn}. Common integration approaches between QNNs and classical networks can be categorized into two types:

One representative work \cite{dutta2025hybrid} incorporates a parameterized quantum circuit as a layer in a classical fully connected neural network, outperforming purely classical models in complex regression and prediction tasks. Similar approaches, such as those in \cite{pandey2hybrid} and \cite{lin2025qaudio}, have integrated QNNs as intermediate layers for deepfake audio detection, achieving exceptional performance. The Q-LoRA proposed in this paper follows this same approach.



Another line of work uses QNNs as final decision layers, following a classical network for feature extraction and fusion. QNNs then produce the final prediction based on the classical network's output. For instance, \cite{xu2026AIGI} employs a QNN as the classification head, achieving superior performance in AI-generated image detection compared to classical algorithms.

\subsection{AI-Generated Content Detection}
With the rapid development of generative models, synthetic content has become increasingly difficult to distinguish from authentic data, making it more prone to misuse in high-risk scenarios such as misinformation dissemination and identity impersonation. AIGC detection aims to determine whether images, audio, and other media are synthesized or manipulated by generative models. This problem is commonly formulated as a binary classification task, where the key objective is to capture residual traces left by the generation process. For images, detectors typically focus on artifacts such as abnormal textures, transform domain \cite{zhang2025towards,karageorgiou2025any}and color distributions \cite{jia2025color}; for audio, they often rely on cues including micro-frequency variations \cite{kumari2025voiceradar} and prosodic patterns to discriminate.

\begin{figure*}[t]                 
  \centering
  \scalebox{0.95}{\includegraphics[width=\textwidth]{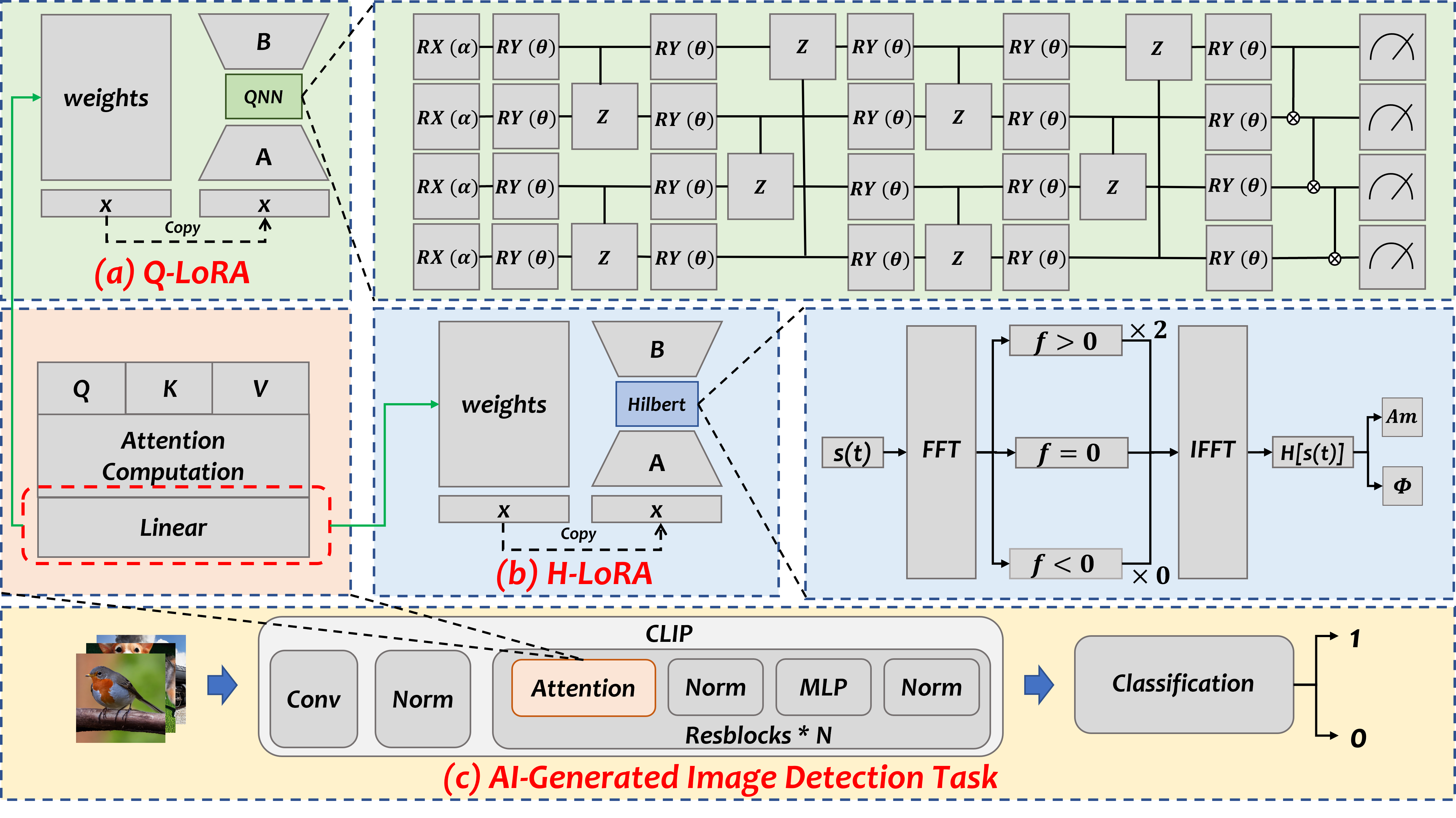}}
  \caption{AI-generated image detection framework: (a) Q-LoRA: A QNN is connected after the CLIP features to extract high-dimensional features through quantum state encoding and processing, and likewise fused with the backbone features via a bypass LoRA module; (b) H-LoRA: The features extracted by CLIP undergo a Hilbert transform to separate and enhance the amplitude and phase characteristics of the signal, thereby enriching the model representation, and are finally fused with the backbone features through a bypass LoRA module, where FFT and IFFT denote Fast Fourier Transform and Inverse Fast Fourier Transform, respectively; (c) Overall Framework:  Illustrates the system pipeline for detecting AI-generated images, which is based on a CLIP visual encoder and integrates multiple efficient fine-tuning adapters.}
  \label{fig:fig_net}
\end{figure*}

\section{Preliminary}

\subsection{LoRA Implementation}

The core idea of LoRA is to achieve parameter-efficient updates through low-rank decomposition. 
Specifically, instead of updating the full weight matrix $W_0 \in \mathbb{R}^{(d\times k)}$, LoRA freezes it and injects a pair of trainable low-rank matrices $B\in \mathbb{R}^{(d\times r)}$ and $A\in \mathbb{R}^{(r\times k)}$, where the rank $r \ll min(d, k)$. The effective weight update is represented as $\Delta W=B \cdot A$. The input feature $x$ is first projected into a low-dimensional subspace via:
\begin{equation}
y=xB^{\mathsf{T}}.
\label{low_proj}
\end{equation}
It is then projected back to the original space through :
\begin{equation}
z=yA^{\mathsf{T}}.
\label{up_proj}
\end{equation}
Finally, the LoRA branch output is scaled by a factor $\alpha /r$ and added to the frozen backbone output:
\begin{equation}
h=W_0x + (\alpha / r) \cdot z.
\label{addori}
\end{equation}
This approach significantly reduces the number of trainable parameters while maintaining the model's expressive capacity, making it suitable for large-scale adaptation scenarios.







\subsection{Hilbert Transform}

Let $s(t)\in \mathbb{R}$ denote a real-valued signal, where $t$ represents a continuous index such as time (for 1D signals) or a spatial coordinate after appropriate vectorization. In the context of neural networks, $s(t)$ can be viewed as a real-valued feature activation along a particular dimension. The Hilbert Transform (HT) is a linear operator $\mathcal{H}[\cdot]$ that maps a real-valued signal to another real-valued signal with a $90^\circ$ phase shift at each frequency component. Formally, it is defined as the convolution:
\begin{equation}
\mathcal{H}[s(t)] = \frac{1}{\pi} \, \text{P.V.} \int_{-\infty}^{\infty} \frac{s(\tau)}{t - \tau} \, d\tau,
\label{hilbert_def}
\end{equation}
where P.V. denotes the Cauchy principal value. In the frequency domain, the HT admits an equivalent and intuitive interpretation: it preserves the magnitude spectrum of $s(t)$ while shifting the phase of positive frequencies by $-\frac{\pi}{2}$ and negative frequencies by $+\frac{\pi}{2}$. Using the Hilbert Transform, one can construct the analytic signal:
\begin{equation}
s_a(t) = s(t) + j \cdot \mathcal{H}[s(t)],
\end{equation}
which embeds the original real-valued signal into a complex-valued representation. This analytic signal admits an explicit amplitude–phase decomposition:
\begin{equation}
A(t) = |s_a(t)|, \quad \Phi(t) = \arg(s_a(t)),
\end{equation}
where $A(t)$ represents the instantaneous amplitude and $\Phi(t)$ represents the instantaneous phase. Importantly, the real and imaginary components of $s_a(t)$ form a quadrature pair: they are orthogonal and jointly encode complementary information about the underlying signal.

In one-dimensional signal processing, the HT is widely used for envelope detection and phase-based analysis in non-stationary signals, with applications in speech, radar, and fault diagnosis. Recent studies leverage HT-based phase representations for detecting synthetic or manipulated audio, demonstrating its effectiveness in forensic settings \cite{shi2025audioHT,unoki2024deepfakeHT}. In two-dimensional image processing, HT and its extensions are applied to spatial feature maps to enable robust edge detection, texture extraction, and phase congruency analysis, which is known to be insensitive to illumination changes \cite{yang2018edge,yang2018hilbert}. HT-based representations are also fundamental in interferometric fringe analysis for 3D shape reconstruction, highlighting their ability to capture fine-grained structural information from complex visual patterns \cite{guerra2023classification}.

\section{Method}
\subsection{Overview}
To investigate whether the empirical benefits of QNNs in few-shot settings stem from their structural inductive biases, namely, phase-aware representations and norm-constrained transformations, we adopt few-shot AIGC detection as a representative benchmark.

As shown in Figure~\ref{fig:fig_net}, our framework builds upon a frozen CLIP visual encoder for feature extraction. The AIGC detection task is formulated as a binary classification problem. To enable efficient adaptation under limited data regimes, we incorporate parameter-efficient adapters in the form of LoRA modules.

Specifically, an input image is first encoded by CLIP into semantic features. LoRA adapters are inserted into selected Transformer layers, where only low-rank updates are learned. Within the LoRA bottleneck, we introduce two alternative designs for structured inductive enhancement: \textbf{Q-LoRA} and \textbf{H-LoRA}. Q-LoRA augments the low-rank subspace with a lightweight quantum neural network, injecting phase-aware and norm-constrained transformations via quantum circuits. H-LoRA retains the same integration strategy but replaces the quantum module with a classical Hilbert transform pipeline that explicitly disentangles amplitude and phase components in feature space. These enhanced features are then projected back to the original space and fused with the backbone activation, followed by a lightweight classifier. Only the adapter and classifier parameters are updated during training.

\subsection{Q-LoRA}
Q-LoRA integrates a QNN module into the LoRA adapter structure, forming a hybrid quantum-classical fine-tuning scheme. The QNN adopts a 4-qubit architecture, as shown in Figure~\ref{fig:fig_net}. Input features are first encoded into quantum states via RX gates. Two entangling layers follow, each consisting of:

\begin{itemize}
  \item Single-qubit RY rotations,
  \item Controlled-Z (CZ) gates applied to qubit pairs (1–2, 3–4) and (2–3, 4–1) for cyclic entanglement.
\end{itemize}

Finally, an additional layer of single-qubit RY rotations is applied, followed by a chain of CNOT gates, thereby introducing global entanglement before all qubits are measured using the Pauli-Z operator to yield the quantum-enhanced feature vector.

This architecture enables:
\begin{itemize}
  \item \textbf{Phase-aware representations}: where information is jointly encoded in orthogonal amplitude and phase components.
  \item \textbf{Norm-constrained transformations}: due to the unitary nature of quantum gates, which regularize the optimization trajectory.
\end{itemize}

Q-LoRA thus enhances feature expressiveness and learning stability while maintaining parameter efficiency through its modular entanglement design.

\subsection{H-LoRA}
We recognize that the core advantages of quantum neural networks may stem from two key structural inductive biases: (i) phase-aware representations, which encode information in orthogonal amplitude–phase components to enrich the expressivity of the feature space; and (ii) norm-constrained structured transformations, which regularize optimization through geometry-preserving transformations. To this end, we propose H-LoRA, a classical Hilbert-transform-based module that mimics quantum-inspired information encoding via analytic signal construction.

The H-LoRA layer functions as a plug-and-play module embedded in the fine-tuning pipeline. It operates on the low-rank feature projection from upstream layers and processes the signal through the following steps:

\paragraph{1) Low-rank projection.} The input feature $x$ is first projected via a low-rank matrix $B$ to obtain a low-dimensional representation $x_l$:
\begin{equation}
x_l = Bx.
\label{low_proj}
\end{equation}

\paragraph{2) Analytic signal construction.} 
To simulate quantum wavefunctions of the form $\psi(t) = A(t)e^{j\phi(t)}$, we construct the analytic signal of $x_l$ via the Hilbert transform:
\begin{equation}
x_a[n] = x_l[n] + j \cdot \mathcal{H}(x_l[n]),
\end{equation}
where $\mathcal{H}(x_l[n])$ is the discrete Hilbert transform. The resulting signal admits an amplitude–phase decomposition:
\begin{equation}
A[n] = |x_a[n]|,\quad \Phi[n] = \arg(x_a[n]).
\end{equation}
These two orthogonal components represent the signal's energy and phase dynamics, analogous to quantum phase–amplitude representations.

\paragraph{3) Phase-aware augmentation.}
We combine the original projected feature $x_l$, the amplitude $A[n]$, and phase $\Phi[n]$ to form the enhanced feature:
\begin{equation}
x_{\text{enhanced}} = x_l + A[n] + \Phi[n].
\end{equation}
This composition models the interaction between real, amplitude, and phase information, mimicking the composite measurement process in quantum systems.

\paragraph{4) Norm-constrained structural transformation.}
Although H-LoRA does not employ unitary matrices, its enhanced features lie in a constrained subspace due to the geometric coupling of $x_l$, $A[n]$, and $\Phi[n]$. Specifically, $A[n]$ and $\Phi[n]$ are derived from $x_l$ and thus maintain bounded norm and correlation:
\[
\|x_{\text{enhanced}}\| \leq \|x_l\| + \|A[n]\| + \|\Phi[n]\|.
\]
This implies a norm-constrained transformation space, which induces a regularizing effect akin to quantum unitaries that satisfy $U^\dagger U = I$.

\paragraph{5) Reprojection and fusion.}
Finally, the enhanced feature is projected back to the original feature dimension using the second low-rank matrix $A$, modulated by a learnable scaling factor $\alpha$:
\begin{align}
\Delta x &= \alpha \cdot A x_{\text{enhanced}} \label{up_proj}, \\
x_{\text{output}} &= x + \Delta x \label{addori}.
\end{align}
This completes the LoRA-style adjustment with Hilbert-based quantum-analogical inductive biases.

\section{Experiments}
\subsection{Settings}



All experiments were performed on the same server platform, equipped with an Intel(R) Xeon(R) Platinum 8369B CPU@2.90GHz as the central processor and an NVIDIA GeForce RTX 4090 GPU with 24GB of video memory as the graphics processor.

\textbf{Image detection task. }We use CLIP as the backbone. Unless otherwise specified, we inject LoRA adapters into the last 6 transformer blocks, with rank $r=4$ and $\alpha=1$, and use a single linear layer as the classifier head. We train for up to 20 epochs with batch size 16 and learning rate $5\times10^{-4}$. The input resolution is $224 \times 224$. Data are sourced from AIGCDetectionBenchMark: for training, we randomly sample from the Stable Diffusion v1.4 subset. For testing, we randomly select 100 images from each generator category: Midjourney, Wukong, VQDM, Stable Diffusion v1.4 (SD1.4), Stable Diffusion v1.5 (SD1.5), Glide, and ADM. We consider few-shot training sizes of 200, 400 and 800, where each number denotes the total number of training examples, including both real and AI-generated images. To assess stability under data scarcity, we run 10 independent trials with seeds $\{0,1,...,9\}$. In each trial, the few-shot training set is re-sampled at random, and we report the average across trials.


\textbf{Audio detection task.} We additionally conduct few-shot audio forgery detection on ASVspoof 2019 LA \cite{todisco2019asvspoof} to examine cross-modality generalization. Whisper \cite{radford2023wisper} serves as the backbone. We inject LoRA adapters into the last 3 encoder layers, with rank $r=4$ and $\alpha=1$, and use a single linear layer as the classifier head. We train for up to 20 epochs with batch size 16 and learning rate $2\times10^{-4}$. Audio is resampled to 16 kHz and cropped or padded to 4 seconds. Since our focus is the few-shot regime, we construct few-shot splits by random sampling: the training and test sets are randomly selected from the full pool (over 25k clips), with 1,500 clips used for testing. As in the image experiments, we repeat 10 trials  with seeds $\{0,1,...,9\}$ and re-sample the few-shot training subset in each trial.


\textbf{Metrics.}  We report Area Under the ROC Curve (AUC), Accuracy (ACC), Precision (PR), and Recall (RE) in few-shot image detection task and report AUC, F1-Score (F1), ACC, Equal Error Rate (EER) in few-shot audio detection task.

\subsection{Results of Few-Shot Image Detection Task} \label{results_5_2}

\begin{table*}[ht!]
  \caption{Performance comparison under different few-shot settings for AI-generated image detection. For each metric, the best and second-best results within the same training sample size are highlighted in bold and underline, respectively.}
  \label{table1}

  \vspace*{0pt}  
  \centering
  \small
  \begin{sc}
    \setlength{\tabcolsep}{2.5pt}
    \renewcommand{\arraystretch}{1.1}
     \begin{tabularx}{\textwidth}{@{}lXXXXXXXXXXXXX@{}}
       \toprule
       \multirow{2}{*}{\makecell{Model \\ SAMPLES}} & \multicolumn{1}{c}{Metrics} & \multicolumn{3}{c}{CLIP} & \multicolumn{3}{c} {LoRA} & 
       \multicolumn{3}{c}{Q-LoRA} & \multicolumn{3}{c}{H-LoRA} \\
       \cmidrule(lr{2-2} \cmidrule(lr){3-5} \cmidrule(lr){6-8} \cmidrule(lr){9-11} \cmidrule(lr){12-14}
        & & 200 & 400 & 800 & 200 & 400 & \multicolumn{1}{c}{800} & 200 & 400 & 800 & 200 & 400 & 800 \\
       \midrule

       \multirow{4}{*}{\textbf{Midjourney}}
       & AUC(\%)  & \underline{98.38}  & \underline{98.29}  & \underline{98.38}  & 94.02  & 95.78  & 97.59  & \textbf{98.62}  & \textbf{98.74}  & \textbf{98.68}  & 96.31  & 97.80  & 97.91 \\
       & ACC(\%)  & \textbf{93.40}  & \textbf{92.85}  & \textbf{91.85}  & 82.60  & 80.55  & 78.70  & \underline{90.40}  & \underline{86.15}  & 81.75  & 86.10  & 81.90  & \underline{84.85}\\
       & PR(\%)   & \textbf{93.86}  & \textbf{90.96}  & \textbf{88.56}  & 77.80  & 73.42  & 71.37  & \underline{87.62}  & \underline{79.93}  & 73.45  & 80.37  & 75.08  & \underline{77.95}\\
       & RE(\%)   & 93.00  & 95.50  & 96.50  & 93.60  & 97.40  & 98.20  & \underline{96.00}  & \textbf{98.60}  & \textbf{99.50}  & \textbf{96.40}  & \underline{98.40}  & \underline{98.70}\\
       \midrule
      
       \multirow{4}{*}{\textbf{Wukong}}
       & AUC(\%)  & 97.31  & 99.00  & 99.27  & 97.22  & 99.34  & 99.85  & \textbf{99.43}  & \underline{99.83}  & \textbf{99.97}  & \underline{99.18}  & \textbf{99.84}  & \underline{99.94} \\
       & ACC(\%)  & 92.70  & 94.95  & 95.90  & 91.00  & 95.30  & 97.80  & \underline{95.00}  & \underline{97.90}  & \textbf{99.12}  & \textbf{95.80}  & \textbf{98.40}   & \underline{98.35}\\
       & PR(\%)   & 93.05  & 94.73  & 95.56  & 88.33  & 93.62  & 96.81  & \underline{94.86}  & \underline{97.60}  & \textbf{98.76}  & \textbf{95.02}  & \textbf{97.86}  & \underline{98.15}\\
       & RE(\%)   & 92.30  & 95.30  & 96.30  & 94.90  & 97.50  & \underline{98.90}  & \underline{95.40}  & \underline{98.30}  & \textbf{99.50}  & \textbf{96.80}  & \textbf{99.00}  & 98.60\\
       \midrule
      
       \multirow{4}{*}{\textbf{SD1.4}}
       & AUC(\%)  & 98.51  & 99.38  & 99.75  & 98.10  & 99.56  & \underline{99.93}  & \underline{99.67}  & \textbf{99.99}  & \textbf{99.99}  & \textbf{99.67}  & \underline{99.84}  & 99.92 \\
       & ACC(\%)  & 94.40  & 96.75  & 97.60  & 93.25  & 96.60  & 98.60  & \underline{96.50}  & \textbf{99.10}  & \textbf{99.75}  & \textbf{97.20}  & \underline{98.75}  & \underline{98.70}\\
       & PR(\%)   & 96.28  & 98.26  & 98.89  & 93.35  & 95.91  & 99.10  & \underline{97.48}  & \underline{99.01}  & \textbf{99.75}  & \textbf{97.93}  & \textbf{99.30}  & \underline{99.51}\\
       & RE(\%)   & 92.40  & 95.20  & 96.30  & 93.50  & 97.60  & \underline{98.10}  & \underline{95.60}  & \textbf{99.20}  & \textbf{99.70}  & \textbf{96.50}  & \underline{98.20}  & 97.90\\
       \midrule
      
       \multirow{4}{*}{\textbf{SD1.5}}
       & AUC(\%)  & 98.74  & 99.32  & 99.72  & 98.61  & 99.54  & 99.86  & \textbf{99.80}  & \textbf{99.98}  & \underline{99.97}  & \underline{99.53}  & \underline{99.93}  & \textbf{99.97} \\
       & ACC(\%)  & 94.70  & 96.35  & 97.25  & 93.55  & 96.75  & 98.05  & \textbf{97.00}  & \textbf{99.15}  & \textbf{99.37}  & \underline{96.90}  & \underline{98.65}  & \underline{98.80}\\
       & PR(\%)   & 96.59  & 97.65  & 97.90  & 93.54  & 96.46  & 98.25  & \textbf{98.80}  & \textbf{99.41}  & \textbf{99.75}  & \underline{98.20}  & \underline{99.19}  & \underline{99.60}\\
       & RE(\%)   & 92.70  & 95.00  & 96.60  & 93.80  & 97.40  & 97.90  & \underline{95.20}  & \textbf{98.90}  & \textbf{99.00}  & \textbf{95.60}  & \underline{98.10}  & \underline{98.00}\\
       \midrule
      
       \multirow{4}{*}{\textbf{Glide}}
       & AUC(\%)  & 94.35  & 97.09  & 98.09  & 92.15  & 96.78  & 99.14  & \textbf{97.60}  & \textbf{99.41}  & \textbf{99.64}  & \underline{97.59}  & \underline{98.99}  & \underline{99.48} \\
       & ACC(\%)  & 85.20  & 89.00  & 90.15  & 80.50  & 86.65  & 91.30  & \underline{89.70}  & \textbf{92.50}  & \textbf{94.87}  & \textbf{90.75}  & \underline{91.95}  & \underline{93.35}\\
       & PR(\%)   & 81.56  & 85.32  & 86.05  & 74.92  & 81.10  & 86.42  & \underline{87.18}  & \underline{88.39}  & \textbf{91.27}  & \textbf{87.03}  & \textbf{87.92}  & \underline{89.42}\\
       & RE(\%)   & 91.30  & 94.60  & 96.10  & 92.70  & 97.50  & 98.60  & \underline{93.60}  & \underline{98.60}  & \textbf{99.50}  & \textbf{96.40}  & \textbf{99.30}  & \underline{99.00}\\
       \midrule
      
       \multirow{4}{*}{\textbf{VQDM}}
       & AUC(\%)  & 91.75  & 95.24  & 96.67  & 91.96  & 96.20  & 98.65  & \underline{95.90}  & \textbf{99.16}  & \underline{99.38}  & \textbf{96.61}  & \underline{98.65}  & \textbf{99.92} \\
       & ACC(\%)  & 81.50  & 84.75  & 86.65  & 81.15  & 84.95  & 88.80  & \underline{85.10}  & \underline{90.00}  & \underline{90.75}  & \textbf{87.75}  & \textbf{91.50}  & \textbf{91.95}\\
       & PR(\%)   & 76.15  & 78.84  & \underline{86.91}  & 76.31  & 79.23  & 82.96  & \underline{79.56}  & \underline{84.76}  & 85.15  & \textbf{82.73}  & \textbf{87.17}  & \textbf{88.12}\\
       & RE(\%)   & 91.90  & 95.20  & 96.10  & 93.20  & 96.50  & \underline{98.20}  & \underline{95.60}  & \underline{98.30}  & \textbf{99.00}  & \textbf{96.40}  & \textbf{98.30}  & 97.70\\
       \midrule

       \multirow{4}{*}{\textbf{ADM}}
       & AUC(\%)  & 87.22  & 91.91  & 93.81  & 80.03  & 87.71  & 94.96  & \textbf{91.33}  & \textbf{96.80}  & \textbf{97.52}  & \underline{90.96}  & \underline{95.34}  & \underline{95.42} \\
       & ACC(\%)  & 73.45  & \underline{77.45}  & \textbf{78.90}  & 68.15  & 70.85  & 74.50  & \underline{74.50}  & 76.00  & 78.12  & \textbf{75.05}  & \textbf{80.35}  & \underline{78.75}\\
       & PR(\%)   & 67.28  & \underline{70.11}  & \textbf{71.38}  & 62.50  & 64.22  & 67.16  & \underline{67.82}  & 68.64  & 69.80  & \textbf{67.97}  & \textbf{73.76}  & \underline{71.09}\\
       & RE(\%)   & 91.60  & 96.10  & 96.80  & 94.40  & 97.00  & 98.10  & 95.20  & \textbf{98.90}  & \textbf{99.30}  & \textbf{96.70}  & \underline{98.40}  & \underline{99.00}\\
       \midrule

       
       
      \multirow{4}{*}{\textbf{Average}}
& \cellcolor{gray!25}AUC(\%)  
& \cellcolor{gray!25}95.18  & \cellcolor{gray!25}97.17  & \cellcolor{gray!25}97.96  
& \cellcolor{gray!25}93.15  & \cellcolor{gray!25}96.42  & \cellcolor{gray!25}98.57  
& \cellcolor{gray!25}\textbf{97.48}  & \cellcolor{gray!25}\textbf{99.13}  & \cellcolor{gray!25}\textbf{99.31}  
& \cellcolor{gray!25}\underline{97.12}  & \cellcolor{gray!25}\underline{98.63}  & \cellcolor{gray!25}\underline{98.74} \\
& \cellcolor{gray!25}ACC(\%)  
& \cellcolor{gray!25}87.91  & \cellcolor{gray!25}90.30  & \cellcolor{gray!25}91.19  
& \cellcolor{gray!25}84.31  & \cellcolor{gray!25}87.38  & \cellcolor{gray!25}89.68  
& \cellcolor{gray!25}\underline{89.75}  & \cellcolor{gray!25}\underline{91.54}  & \cellcolor{gray!25}\underline{91.96}  
& \cellcolor{gray!25}\textbf{89.94}  & \cellcolor{gray!25}\textbf{91.64}  & \cellcolor{gray!25}\textbf{92.11} \\
& \cellcolor{gray!25}PR(\%)  
& \cellcolor{gray!25}86.40  & \cellcolor{gray!25}87.98  & \cellcolor{gray!25}\underline{88.64}  
& \cellcolor{gray!25}80.96  & \cellcolor{gray!25}83.42  & \cellcolor{gray!25}86.01  
& \cellcolor{gray!25}\textbf{87.61}  & \cellcolor{gray!25}\underline{88.25}  & \cellcolor{gray!25}88.28  
& \cellcolor{gray!25}\underline{87.03}  & \cellcolor{gray!25}\textbf{88.61}  & \cellcolor{gray!25}\textbf{89.12} \\
& \cellcolor{gray!25}RE(\%)  
& \cellcolor{gray!25}92.17  & \cellcolor{gray!25}95.27  & \cellcolor{gray!25}96.39  
& \cellcolor{gray!25}93.76  & \cellcolor{gray!25}97.27  & \cellcolor{gray!25}98.10  
& \cellcolor{gray!25}\underline{95.29}  & \cellcolor{gray!25}\textbf{98.69}  & \cellcolor{gray!25}\textbf{99.38}  
& \cellcolor{gray!25}\textbf{96.40}  & \cellcolor{gray!25}\underline{98.53}  & \cellcolor{gray!25}\underline{98.44} \\
       
     \bottomrule
     \end{tabularx}
  \end{sc}
  \vskip -0.1in
\end{table*}

\textbf{Quantitative analysis.} In this section, we perform experiments to evaluate these methods under different few-shot settings for AI-generated image detection. Specifically, we use (i) CLIP with only the classification head fine-tuned, (ii) standard LoRA, (iii) Q-LoRA, and (iv) our classical simulation method H-LoRA.

Table \ref{table1} summarizes cross-generator generalization when training on SD1.4 and testing on diverse generators under few-shot settings. Q-LoRA and H-LoRA consistently outperform standard LoRA in the most constrained regime (200 samples), with Q-LoRA achieving an average ACC of 89.75\% (an improvement of 5.44\% over LoRA) and H-LoRA achieving 89.94\% (an improvement of 5.63\% over LoRA). As the training size increases to 400 and 800 samples, performance improves for all methods, and the relative gaps narrow.

Across most test generators, the improvements of Q-LoRA and H-LoRA are not driven by a single category, but are more pronounced on generators farther from the training distribution, where standard LoRA is more prone to overfitting SD1.4-specific artifacts. The unweighted averages in Table~\ref{table1} indicate that H-LoRA remains consistently competitive with Q-LoRA across training sizes and metrics, while both outperform standard LoRA. The simultaneous gains in threshold-free AUC and threshold-dependent ACC suggest that the improvements reflect more separable representations for real versus generated content, rather than calibration effects. These results indicate that H-LoRA reproduces much of Q-LoRA’s benefit using a purely classical implementation.

\begin{figure}[h]                 
  \centering
  \includegraphics[width=\linewidth]{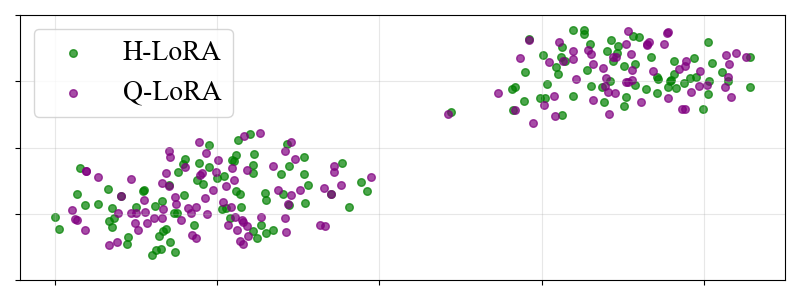}
  \caption{Q-LoRA and H-LoRA t-SNE visualization}
  \label{fig:tsne}
\end{figure}

\begin{figure}[h]
\centering
\begin{subfigure}{0.30\linewidth}
  \centering
  \includegraphics[width=\linewidth]{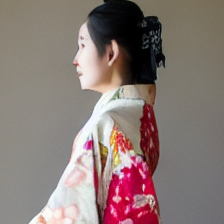}
\end{subfigure}\hfill
\begin{subfigure}{0.30\linewidth}
  \centering
  \includegraphics[width=\linewidth]{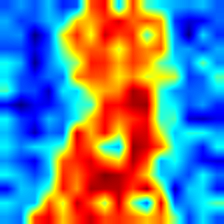}
\end{subfigure}\hfill
\begin{subfigure}{0.30\linewidth}
  \centering
  \includegraphics[width=\linewidth]{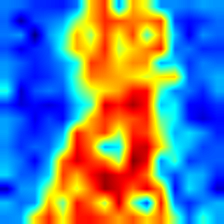}
\end{subfigure}


\begin{subfigure}{0.30\linewidth}
  \centering
  \includegraphics[width=\linewidth]{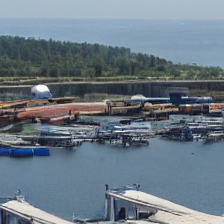}
  \caption{Original Image}
\end{subfigure}\hfill
\begin{subfigure}{0.30\linewidth}
  \centering
  \includegraphics[width=\linewidth]{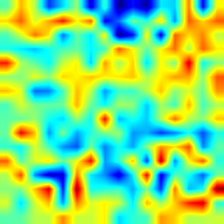}
  \caption{Q-LoRA}
\end{subfigure}\hfill
\begin{subfigure}{0.30\linewidth}
  \centering
  \includegraphics[width=\linewidth]{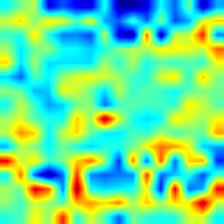}
  \caption{H-LoRA}
\end{subfigure}
\caption{Attention visualization of Q-LoRA and H-LoRA}
\label{fig:Attention_Vis}
\end{figure}

\textbf{Visualization analysis.} Beyond aggregate metrics, we also show the visual results of the similarity of the feature extracted by H-LoRA and Q-LoRA. 
Figure \ref{fig:tsne} shows a t-SNE projection of the learned features, where the embeddings produced by H-LoRA and Q-LoRA exhibit substantial overlap in the projected space, suggesting that the two methods induce broadly similar feature distributions under t-SNE projection. Figure \ref{fig:Attention_Vis} further visualizes attention maps on representative examples; the spatial patterns highlighted by H-LoRA are highly consistent with those of Q-LoRA, indicating that the two adaptations attend to comparable regions that are informative for detection. Together, these visualizations complement the quantitative results by illustrating that the proposed classical surrogate can largely reproduce key representation characteristics associated with the quantum-inspired variant.


\subsection{Results of Few-Shot Audio Detection Task}
Besides image detection, we also perform the generated audio detection experiments.
We compare (i) Whisper with only the classification head fine-tuned, (ii) standard LoRA, and (iii) our classical simulation method H-LoRA under different few-shot settings for audio forgery detection. As shown in Table \ref{table2}, under random few-shot splits, H-LoRA achieves consistent improvements over Whisper with linear classifier probing and standard LoRA across 50, 100 and 200-shot. The gains are reflected not only in the best in threshold-free AUC (99.43\%) but also in in the highest threshold-dependent metrics, such as F1 (96.73\%) and ACC (96.77\%), and are accompanied by a lower EER (0.0333), which is a primary metric in ASVspoof-style anti-spoofing evaluation. Furthermore, under the 50-sample setting, H-LoRA' s 90.69\% demonstrates a more significant advantage over LoRA's 72.99\% in terms of ACC. These trends suggest that the proposed phase-structured reparameterization provides a useful inductive bias beyond the image domain.

\begin{table}[H]
  \caption{Algorithm performance with varying sample sizes in audio forgery detection}
  \label{table2}
  \vskip 0.05in
  \begin{center}
    \begin{small}
      \begin{sc}
      \setlength{\tabcolsep}{3pt}              
        \begin{tabular}{lccccc}
          \toprule
          Method       & Samples   & {AUC(\%)}     & {F1(\%)}          & {ACC(\%)}       & {EER}  \\
          \midrule
                       & 50        & 90.85           & 72.54           & 72.99           & 0.1616 \\
          Whisper       & 100       & 93.71           & 87.42           & 88.19           & 0.1260 \\
                       & 200       & 95.82           & 89.93           & 90.69           & 0.0957 \\
          \midrule
                       & 50        & 92.86           & 78.89           & 80.34           & 0.1358 \\
          LoRA         & 100       & 97.45           & 89.39           & 90.40           & 0.0711 \\
                       & 200       & 99.04           & 95.54           & 95.67           & 0.0431 \\
         \midrule
                       & 50        & 96.52           & 88.32           & 89.33           & 0.0871 \\
          H-LoRA       & 100       & 98.57           & 94.37           & 94.58           & 0.0548 \\
                       & 200       & \textbf{99.43}  & \textbf{96.73}  & \textbf{96.77}  & \textbf{0.0333} \\
          \bottomrule
        \end{tabular}
      \end{sc}
    \end{small}
  \end{center}
  \vskip -0.1in
\end{table}

\subsection{Efficiency and Practicality Analysis}

\begin{table}[H]
  \caption{Efficiency comparison between H-LoRA and Q-LoRA.}
  \label{table_speed}
  \vskip 0.05in
  \begin{center}
    \begin{small}
      \begin{sc}
        \begin{tabular}{lcc}
          \toprule
          Method    & {H-LoRA}      & {Q-LoRA}  \\
          \midrule                    
          Inference Time(sec)    & \textbf{0.09}    & 65.68     \\   
          Training Time(epoch/sec)    & \textbf{4.07}     & 2088.34 \\
          Additional Parameter  & \textbf{0}           & 24\\
        \bottomrule
        \end{tabular}
      \end{sc}
    \end{small}
  \end{center}
  \vskip -0.1in
\end{table}

In this section, we compare the efficiency H-LoRA and Q-LoRA for image detection task. As seen in Table \ref{table_speed}, despite achieving comparable detection performance, H-LoRA demonstrates overwhelming advantages in both training and inference efficiency. Specifically, H-LoRA requires only 0.09 seconds for inference and 4.07 seconds per training epoch, whereas Q-LoRA incurs 65.68 seconds for inference and over 2,000 seconds per epoch. Moreover, H-LoRA compared to LoRA introduces no additional trainable parameters, in contrast to Q-LoRA, which adds 24 parameters. These results clearly indicate that H-LoRA effectively preserves the efficiency of standard LoRA while avoiding the substantial time overhead introduced by explicit QNN.

\subsection{Ablation Experiment}

\begin{figure}[h]
\centering
\begin{subfigure}{0.49\linewidth}
  \centering
  \includegraphics[width=\linewidth]{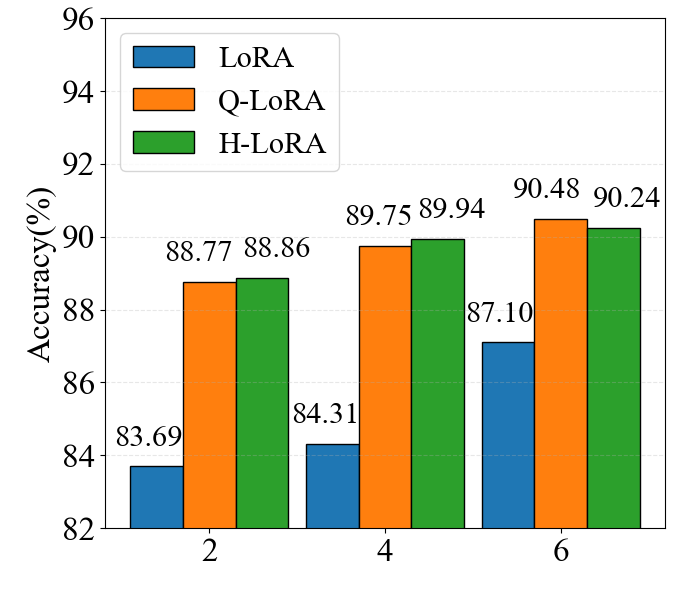}
  \caption{Rank}
  \label{fig:Different_Rank}
\end{subfigure}\hfill
\begin{subfigure}{0.49\linewidth}
  \centering
  \includegraphics[width=\linewidth]{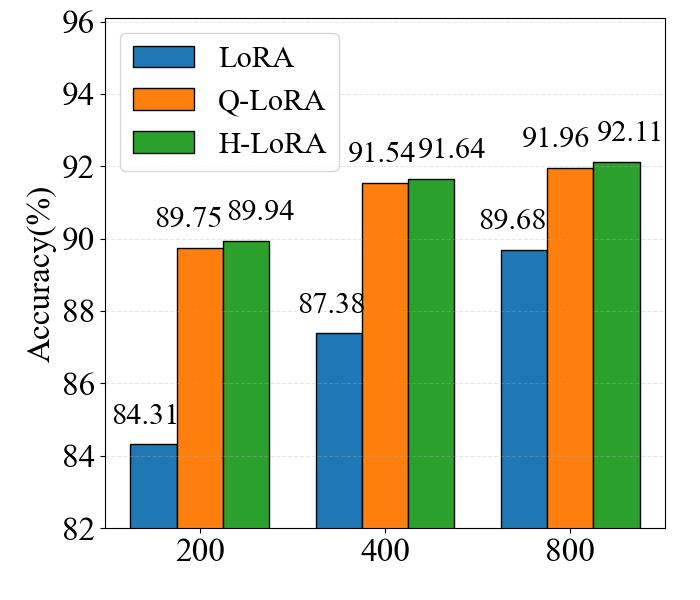}
  \caption{Samples}
  \label{fig:Different_Training_Samples}
\end{subfigure}

\caption{Accuracy of LoRA, Q-LoRA and H-LoRA under different ranks and training sample sizes}
\end{figure}

We conduct ablations to better evaluate which factors contribute to the improvements of H-LoRA.

\textbf{Effect of rank.} We vary the adapter rank $r \in \{2, 4, 6\}$ while keeping the rest of the training protocol fixed, and test the detection accuracy. As shown in the figure \ref{fig:Different_Rank}, increasing rank improves performance for all methods, reflecting the increased expressive capacity of higher-rank adaptations. However, H-LoRA and Q-LoRA consistently outperform standard LoRA across all rank settings. The relative gains are especially evident at $r=2$ where we observe that the proposed structured transformation is beneficial even under a severely constrained parameter budget.


\textbf{Effect of training sample size.} The ablation results in Figure \ref{fig:Different_Training_Samples} are consistent with the observations in Section \ref{results_5_2}: Q-LoRA and H-LoRA yield the largest performance gains in the lowest-data regime, while the gap narrows as the training sample size increases, indicating reduced reliance on strong inductive bias when more supervision is available.

\begin{table}[H]
  \caption{Performance of different types of intermediate layers in LoRA}
  \label{table_layers}
  \vskip 0.05in
  \begin{center}
    \begin{small}
      \begin{sc}
      \setlength{\tabcolsep}{4.5pt}              
        \begin{tabular}{lccccc}
          \toprule
          Method      & {H-LoRA}      & {Linear}   & {Tanh}     & {Sigmoid}    & {Silu}   \\
          \midrule                    
          AUC(\%)     & \textbf{97.12}         & 96.32      & 96.51      & 96.89        & 97.01    \\ 
          ACC(\%)     & \textbf{89.94}         & 87.79      & 88.30      & 87.19        & 88.36    \\
          Pr(\%)      & \textbf{87.03}         & 85.57      & 85.70      & 84.20     & 85.91     \\
          Re(\%)      & \textbf{96.40}         & 96.44      & 95.14      & 95.87        & 95.59    \\
        \bottomrule
        \end{tabular}
      \end{sc}
    \end{small}
  \end{center}
  \vskip -0.1in
\end{table}

\textbf{Effect of nonlinearity.} To test whether the gains stem merely from added nonlinearity on low rank features, we replace the intermediate transformation in the LoRA pathway with common activation functions or linear layers and evaluate the resulting variants. The results in Table \ref{table_layers} show that these generic nonlinear or linear insertions do not match the performance of H-LoRA in terms of ACC, with all achieving values below 89\%, despite introducing additional expressive components. Moreover, Table \ref{table_n_layers} shows that increasing the number of intermediate Linear layers yields diminishing returns: performance tends to plateau around 89.5\%, and ACC does not improve monotonically with more layers. Together, these findings support the view that the improvement is not merely due to adding nonlinearity, but is more closely related to the proposed phase-aware representation and norm-constrained structured transformation.

\begin{table}[H]
  \caption{Performance of Number of Intermediate Linear Layers in LoRA}
  \label{table_n_layers}
  \vskip 0.05in
  \begin{center}
    \begin{small}
      \begin{sc}
      \setlength{\tabcolsep}{3.5pt}              
        \begin{tabular}{lcccccc}
          \toprule
          Method    & {H-LoRA}      & {1}   & {2}     & {3}    & {4}   & {5}\\
          \midrule                    
          AUC(\%)    & 97.12    & 96.32      & 97.39      & 97.52        & 97.58    & \textbf{97.61}\\   
          ACC(\%)    & \textbf{89.94}     & 87.79      & 88.90      & 89.60        & 89.10    & 89.30\\
          Pr(\%)     & \textbf{87.03}     & 85.57      & 86.04      & 86.25        & 85.77    & 86.68 \\
          Re(\%)     & 96.40      & 96.44      & 96.41      & \textbf{97.19}        & 97.17    & 96.23\\
        \bottomrule
        \end{tabular}
      \end{sc}
    \end{small}
  \end{center}
  \vskip -0.1in
\end{table}

\section{Conclusion}
In this work, we introduce H-LoRA, a fully classical yet quantum-inspired fine-tuning method for few-shot learning. Starting from the empirical success of Q-LoRA, we analyze the structural inductive biases behind its performance, specifically, phase-aware representations and norm-constrained transformations, and show that these can be reproduced classically using the Hilbert transform. By integrating this insight into the LoRA adapter, H-LoRA retains the generalization benefits of Q-LoRA without incurring its heavy simulation cost. Experimental results on few-shot AIGC detection confirm that H-LoRA matches or exceeds Q-LoRA’s accuracy, while offering significantly improved efficiency. 



\section*{Acknowledgements}

This work was supported in part by the National Natural Science Foundation of China under Grant 62272003, in part by the Quantum Science and Technology-National Science and Technology Major Project under Grant 2021ZD0302300, in part by the Science and Technology Major Project of Anhui Province under Grant 202423s06050001.

\section*{Impact Statement}


This paper presents work whose goal is to advance the field of Machine
Learning. There are many potential societal consequences of our work, none
which we feel must be specifically highlighted here.


\nocite{langley00}

\bibliography{example_paper}
\bibliographystyle{icml2026}

\newpage
\appendix
\onecolumn

\section{Hilbert Transform of Discrete Signals}

Given a discrete signal \( x(n) \) of length \( N \), where \( n = 0, 1, 2, \dots, N-1 \), the \( N \)-point Discrete Fourier Transform (DFT) of \( x(n) \) is computed to obtain its frequency-domain representation \( X(k) \), where \( k = 0, 1, 2, \dots, N-1 \):

\[
X(k) = \sum_{n=0}^{N-1} x(n) \cdot \exp\left( -j \cdot \frac{2\pi k n}{N} \right)
\]

Next, the negative frequency components of \( X(k) \) (i.e., for \( k > \frac{N}{2} \)) are inverted in magnitude to obtain the symmetric frequency-domain representation \( Y(k) \):

\[
Y(k) = 
\begin{cases} 
X(k) & \text{for } k = 0, 1, \dots, \frac{N}{2}, \\
-X(k) & \text{for } k = \frac{N}{2}+1, \frac{N}{2}+2, \dots, N-1.
\end{cases}
\]

Finally, the inverse \( N \)-point Discrete Fourier Transform (IDFT) of the symmetric frequency-domain representation \( Y(k) \) is applied to obtain the Hilbert-transformed discrete signal \( h(n) \):

\[
h(n) = \frac{1}{N} \sum_{k=0}^{N-1} Y(k) \cdot \exp\left( j \cdot \frac{2\pi k n}{N} \right)
\]

This results in the Hilbert transform of the original discrete signal \( x(n) \).

\section{More Attention Visualization}
\begin{figure}[h]
\begin{subfigure}{0.15\linewidth}
  \centering
  \includegraphics[width=\linewidth]{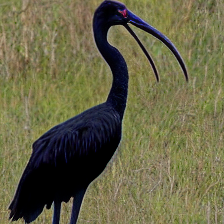}
\end{subfigure}\hfill
\begin{subfigure}{0.15\linewidth}
  \centering
  \includegraphics[width=\linewidth]{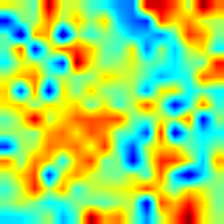}
\end{subfigure}\hfill
\begin{subfigure}{0.15\linewidth}
  \centering
  \includegraphics[width=\linewidth]{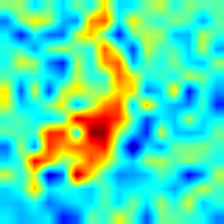}
\end{subfigure}\hfill
\begin{subfigure}{0.15\linewidth}
  \centering
  \includegraphics[width=\linewidth]{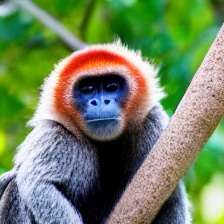}
\end{subfigure}\hfill
\begin{subfigure}{0.15\linewidth}
  \centering
  \includegraphics[width=\linewidth]{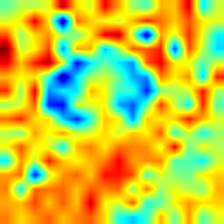}
\end{subfigure}\hfill
\begin{subfigure}{0.15\linewidth}
  \centering
  \includegraphics[width=\linewidth]{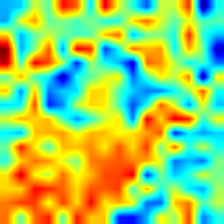}
\end{subfigure}

\begin{subfigure}{0.15\linewidth}
  \centering
  \captionsetup{labelformat=empty}
  \includegraphics[width=\linewidth]{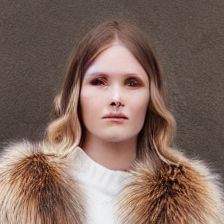}
  \caption{Original Image}
\end{subfigure}\hfill
\begin{subfigure}{0.15\linewidth}
  \centering
  \captionsetup{labelformat=empty}
  \includegraphics[width=\linewidth]{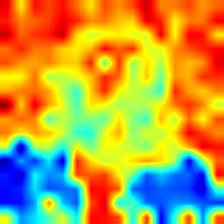}
  \caption{Q-LoRA}
\end{subfigure}\hfill
\begin{subfigure}{0.15\linewidth}
  \centering
  \captionsetup{labelformat=empty}
  \includegraphics[width=\linewidth]{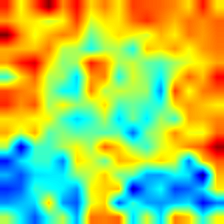}
  \caption{H-LoRA}
\end{subfigure}\hfill
\begin{subfigure}{0.15\linewidth}
  \centering
  \captionsetup{labelformat=empty}
  \includegraphics[width=\linewidth]{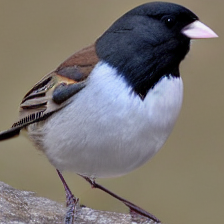}
  \caption{Original Image}
\end{subfigure}\hfill
\begin{subfigure}{0.15\linewidth}
  \centering
  \captionsetup{labelformat=empty}
  \includegraphics[width=\linewidth]{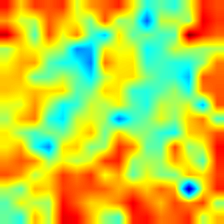}
  \caption{Q-LoRA}
\end{subfigure}\hfill
\begin{subfigure}{0.15\linewidth}
  \centering
  \captionsetup{labelformat=empty}
  \includegraphics[width=\linewidth]{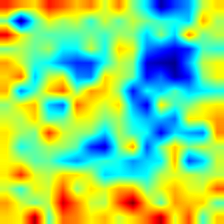}
  \caption{H-LoRA}
\end{subfigure}

\end{figure}

We further conducted additional visualization experiments to demonstrate the similarity between H-LoRA and Q-LoRA in feature capturing. The images on the left show the original images, and the ones on the right show the corresponding attention heatmaps. As can be seen, both H-LoRA and Q-LoRA focus on key areas of the images (such as the prominent parts of the objects) with high similarity. This indicates that the two methods exhibit consistent performance in feature capturing, providing further support for the two possible structural inductive biases.

\section{Training Loss Curve of H-LoRA}

We present the training loss curves for both the image and audio experiments. As shown in the left figure, the image experiment demonstrates a rapid convergence, indicating that the model is easy to train. Similarly, the right figure, showing the audio experiment, also reveals a relatively smooth and fast convergence, further validating the ease of training with H-LoRA across different domains. These results suggest that H-LoRA efficiently learns the task, even in low-data regimes.

\begin{figure}[!t]
\vspace*{-\topskip}
\begin{subfigure}{0.5\linewidth}
  \centering
  \includegraphics[width=\linewidth]{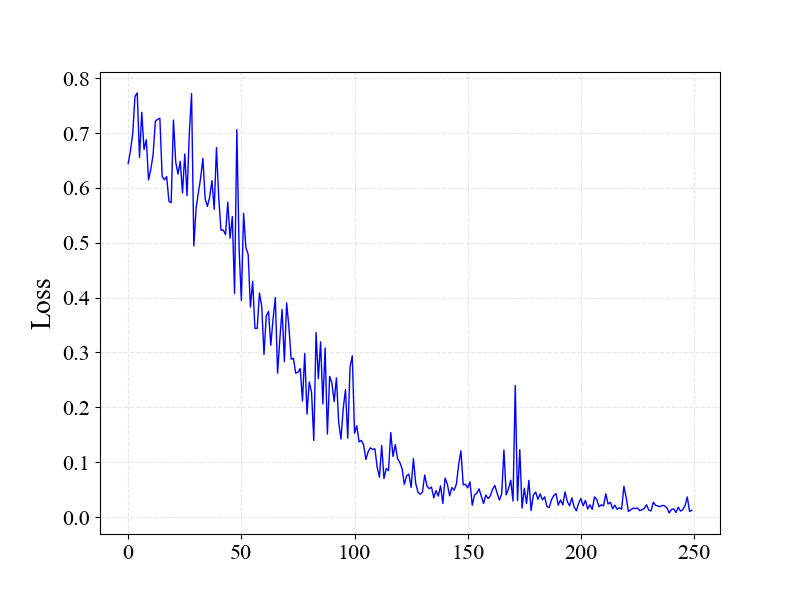}
  \caption{Training Loss of H-LoRA in image detection task}
\end{subfigure}\hfill
\begin{subfigure}{0.5\linewidth}
  \centering
  \includegraphics[width=\linewidth]{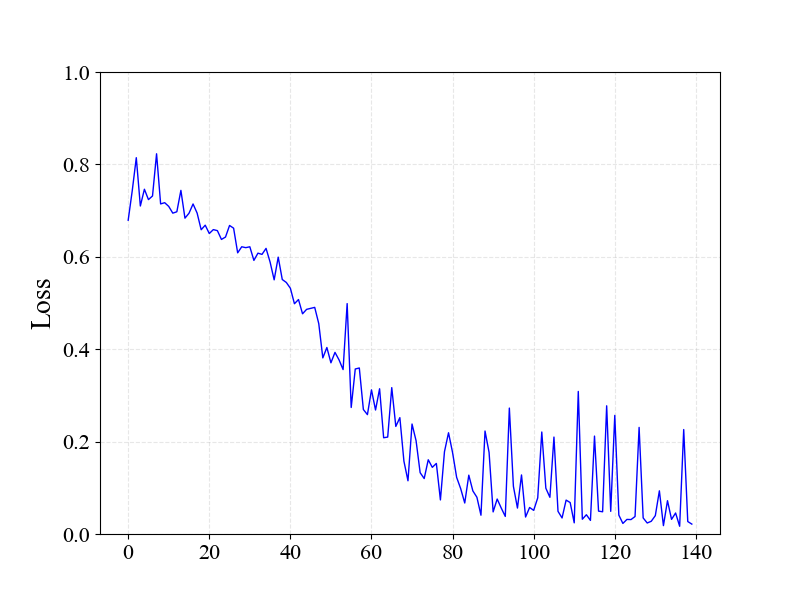}
  \caption{Training Loss of H-LoRA in audio detection task}
\end{subfigure}
\end{figure}



\end{document}